\documentclass[letterpaper, 10 pt, journal, twoside]{IEEEtran} 

\IEEEoverridecommandlockouts                              

\usepackage{cite}
\usepackage{subfigure}
\usepackage{booktabs}
\usepackage{balance}
\usepackage{color}
\usepackage{tabularx}
\usepackage{graphicx,dblfloatfix} 
\usepackage{epsfig} 
\usepackage{amsmath}
\usepackage{amssymb}
\usepackage{listings}
\usepackage{textcomp}
\usepackage{url}
\usepackage{multirow}
\usepackage{todonotes}
\usepackage{hyperref}
\usepackage{cleveref}
\usepackage{wrapfig}
\crefformat{footnote}{#2\footnotemark[#1]#3}
\usepackage{ragged2e} 
\usepackage[subnum]{cases}
\newcolumntype{Y}{>{\RaggedRight\arraybackslash}X} 
\usepackage{algorithm}
\usepackage[noend]{algpseudocode}

\usepackage{tikz}
\usepackage{tkz-euclide}
\usetikzlibrary{math}
\usepackage{pgfplots}
\usetikzlibrary{arrows}

\usepackage{float}

\newcommand{\degree}{$^{\circ}$}

\title{Walking Posture Adaptation for Legged Robot Navigation in Confined Spaces}

\author{Russell Buchanan$^{1}$, Tirthankar Bandyopadhyay$^{2}$, Marko Bjelonic$^{3}$,\\ Lorenz Wellhausen$^{3}$, Marco Hutter$^{3}$, and Navinda Kottege$^{2}$%
\thanks{Manuscript received: September, 10, 2018; Revised December, 23, 2018; Accepted January, 23, 2019.This paper was recommended for publication by Editor N. Tsagarakis upon evaluation of the Associate Editor and Reviewers' comments.
This work was supported by funding from the Commonwealth Scientific and Industrial Research Organisation (CSIRO). \textit{(Corresponding author: Navinda Kottege {\tt\footnotesize navinda.kottege@csiro.au})}} 
\thanks{$^{1}$ R. Buchanan was a student at the Robotic Systems Lab, ETH Z\"urich, 8092 Z\"urich, Switzerland and an intern at the Robotics and Autonomous Systems Group, CSIRO, Pullenvale, QLD 4069, Australia at the time of this work.}%
\thanks{$^{2}$T. Bandyopadhyay and N. Kottege are with Robotics and Autonomous Systems Group, CSIRO, Pullenvale, QLD 4069, Australia.}%
\thanks{$^{3}$M. Hutter, M. Bjelonic and L. Wellhausen are with the Robotic Systems Lab, ETH Z\"urich, 8092 Z\"urich, Switzerland. }%
\thanks{\copyright2019 IEEE.  Personal use of this material is permitted.  Permission from IEEE must be obtained for all other uses, in any current or future media, including reprinting/republishing this material for advertising or promotional purposes, creating new collective works, for resale or redistribution to servers or lists, or reuse of any copyrighted component of this work in other works.}%
\thanks{Digital Object Identifier (DOI): see top of this page.}
}

\begin{document}
\maketitle
\markboth{IEEE Robotics and Automation Letters. Preprint Version. Accepted January, 2019}
{Buchanan \MakeLowercase{\textit{et al.}}: Walking Posture Adaptation for Legged Robots} 

\begin{abstract}

Legged robots have the ability to adapt their walking posture to navigate confined spaces due to their high degrees of freedom. However, this has not been exploited in most common multilegged platforms. This paper presents a deformable bounding box abstraction of the robot model, with accompanying mapping and planning strategies, that enable a legged robot to autonomously change its body shape to navigate confined spaces. The mapping is achieved using robot-centric multi-elevation maps generated with distance sensors carried by the robot. The path planning is based on the trajectory optimisation algorithm CHOMP which creates smooth trajectories while avoiding obstacles. The proposed method has been tested in simulation and implemented on the hexapod robot Weaver, which is 33\,cm tall and 82\,cm wide when walking normally. We demonstrate navigating under 25\,cm overhanging obstacles, through 70\,cm wide gaps and over 22\,cm high obstacles in both artificial testing spaces and realistic environments, including a subterranean mining tunnel.
\end{abstract}

\section{Introduction}
\label{sec:introduction}
\IEEEPARstart{M}{ultilegged} robots are well suited for complex, rough and unstructured terrain. Their many degrees of freedom (DOF) enable navigation of challenging environments including confined spaces. Hexapod robots such as Weaver~\cite{bjelonic_weaver:_nodate} and Lauron V~\cite{roennau_lauron_2014} are very stable statically and capable of walking on rough terrain and up steep inclines. MAX~\cite{elfes_multilegged_2017}, an Ultralight Legged Robot (ULR) was designed to maximise locomotion efficiency in challenging outdoor environments. More agile and efficient, quadrupedal robots such as HyQ2Max \cite{semini_design_2017},  ANYmal~\cite{hutter_anymal_2016}, Minitaur~\cite{kenneally2016design} are capable of running and jumping.

Several robots have shown the ability to change their posture while walking in confined spaces  such as SpotMini from Boston Dynamics (see video~\cite{bd_youtube_2016}), ANYmal~(see video at 24\,s~\cite{anymal_youtube_2016}) and the magnetic foot climbing robot Magneto~\cite{bandyopadhyay_magneto}; however, to the best of our knowledge, none of these posture adaptations are automatically planned.

To autonomously navigate difficult environments, legged robots require mapping and planning techniques together with appropriate control. Weaver, for example, uses proprioceptive terrain characterisation and an admittance controller~\cite{bjelonic_autonomous_2017} for uneven terrain. ANYmal uses robot-centric elevation mapping to select individual footholds that are both on suitable terrain and satisfy kinematic constraints~\cite{fankhauser_robust_2018}. However, neither of these methods adapt the robot's posture for walking through confined environments.

Despite the capability of legged robots to change their walking posture, there exist no solutions to achieve this autonomously. To address this lack of perception and planning methods, we present a solution that enables autonomous navigation of confined spaces by multilegged robots.

\begin{figure}[t!]
    \centering
    \includegraphics[width=70mm]{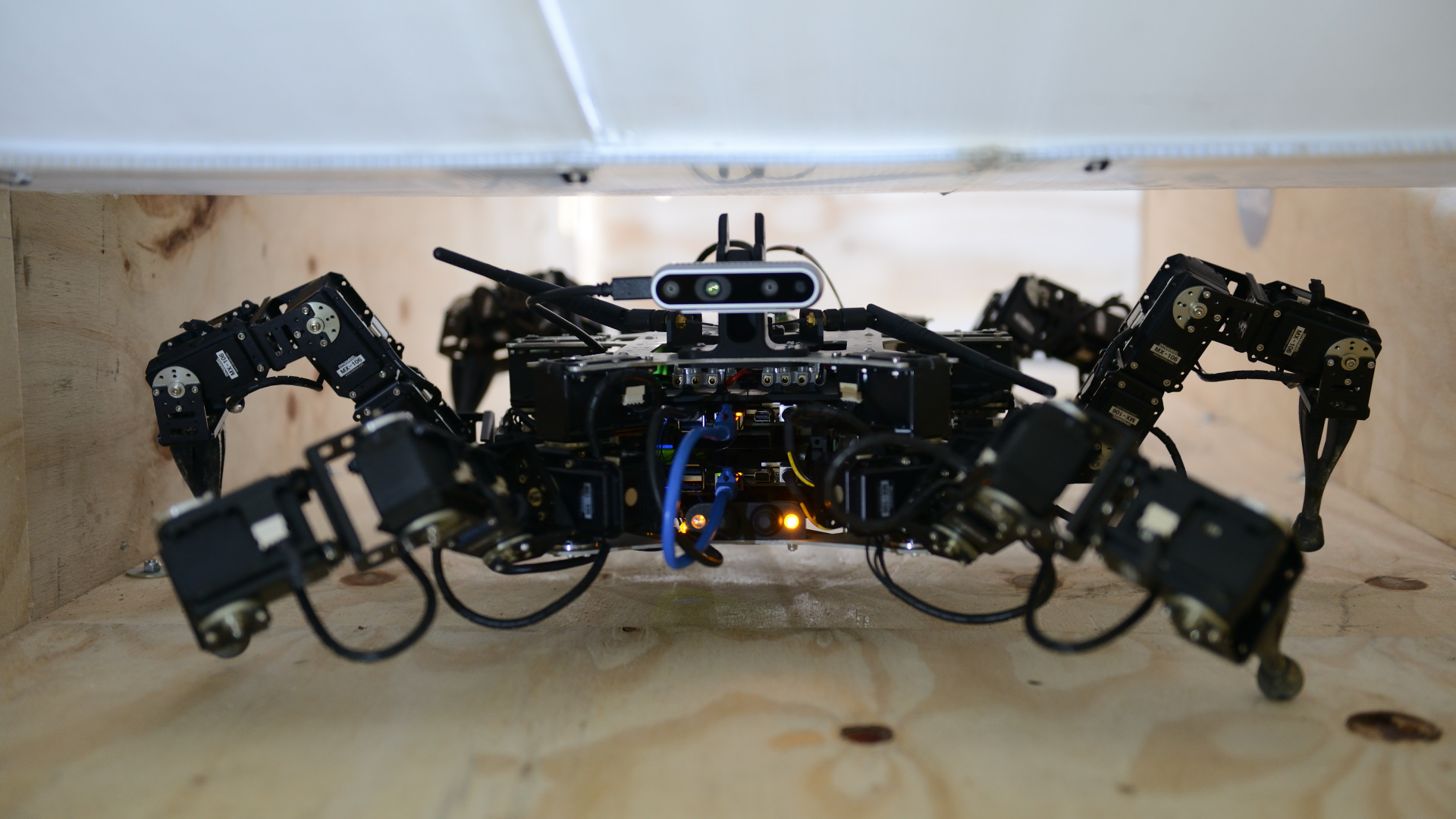}
    \caption{The hexapod robot Weaver adapting its walking posture to pass under an overhanging obstacle with 25\,cm clearance.}
    \label{fig:robot_low}
    \vspace{-0.5cm}
\end{figure}

\subsection{Related Work}
There has been significant progress in navigation of rough terrain for multilegged robots. Terrain characterisation for gait adaptation has been shown to improve locomotion efficiency when walking on rough terrain~\cite{bjelonic_autonomous_2017}. Elevation maps have been used with characterisation to plan footholds for optimal stability and obstacle avoidance~\cite{belter_employing_2018}, as well as walking over gaps and climbing stairs~\cite{fankhauser_robust_2018}. Optimising robot dynamics can enable even more dynamic motions~\cite{winkler_gait_2018}; however, none of these methods consider collisions with the body of the robot, which is necessary in confined spaces.

For complex 3D environments such as large steps, trusses or vehicle egress, full body contact planners using random sampling such as probabilistic roadmaps (PRM) are used \cite{hauser_multi-modal_2010,tonneau_efficient_2018,short_legged_2018}. This can be done by randomly sampling a subspace of all possible contacts limited by stability and reachability as in Tonneau et al.~\cite{tonneau_efficient_2018}. However, this method requires accurate knowledge of the environment in advance. Short and Bandyopadhyay~\cite{short_legged_2018} deal with this by first pre-computing a set of possible configurations based on the robot model assuming no obstacles then selecting the best configurations for a given, dynamic environment. The computational cost of these planners are highly dependant on the complexity of the terrain and the number of joints of the robot. This makes them difficult to apply to high DOF, multilegged robots.

A common strategy is to simplify the problem by using a lower dimensional abstraction of the robot model and planning for this instead. Grey et al.~\cite{grey_footstep} use bounding boxes attached to the robot body to limit the search space of possible configurations. In confined spaces, the bounding boxes are too large and the planner must fall back to using the robot's minimal geometry. Orthney et al.~\cite{orthey_quotient} deal with confined spaces by using nested robots to find paths for progressively less abstract models.

\subsection{Contributions}

We take inspiration from soft robotics and propose a deformable bounding box abstraction of the robot model. The concept is similar to~\cite{grey_footstep}, however, our bounding box can change in volume, allowing it to continue to be effective in confined spaces. Unlike~\cite{orthey_quotient}, as our abstraction changes, the complexity of the planning problem remains the same which allows for fast computation in confined spaces. We do not plan the foot tip locations, however our method could later be used with a leg swing planner such as in \cite{fankhauser_robust_2018} to avoid any collision of the robot's legs with the terrain.

Paths for deformable robots can be planned with sampling methods such as PRM \cite{burchan_probabilistic_nodate}. Yoshida et al.~\cite{yoshida_simulation-based_2015} use a modified covariant trajectory optimisation method based on CHOMP~\cite{zucker_chomp:_2013} to plan trajectories for an elastic O-ring. These approaches work with soft, deformable objects and use methods such as Free-Form Deformation (FFD) or Finite Element Method (FEM) to calculate how the robot's body should deform under pressure from the environment. As a result, these methods cannot be directly applied to rigid multi-joint robots.
 
We plan body posture trajectories for the proposed deformable bounding box abstraction of a multilegged robot. Collision checking is done in a signed distance field (SDF) which is generated from a robot-centric multi-elevation map. For the planner, we employ CHOMP~\cite{zucker_chomp:_2013} although any planner could be used. The contributions from this work can be listed as follows:

\begin{itemize}
    \item Introduction of a deformable bounding box abstraction of the robot model.
    \item Present a planning framework for the deformations and demonstrate how trajectory optimisation is applied.
    \item Demonstration of posture adaptation on a real robot in various real scenarios.
    \item Extension of robot-centric elevation mapping to full 3D space mapping.
\end{itemize}


\section{Deformable Abstraction}
\label{sec:definition} 

\begin{figure}[t!]
    \centering
    \includegraphics[width=65mm]{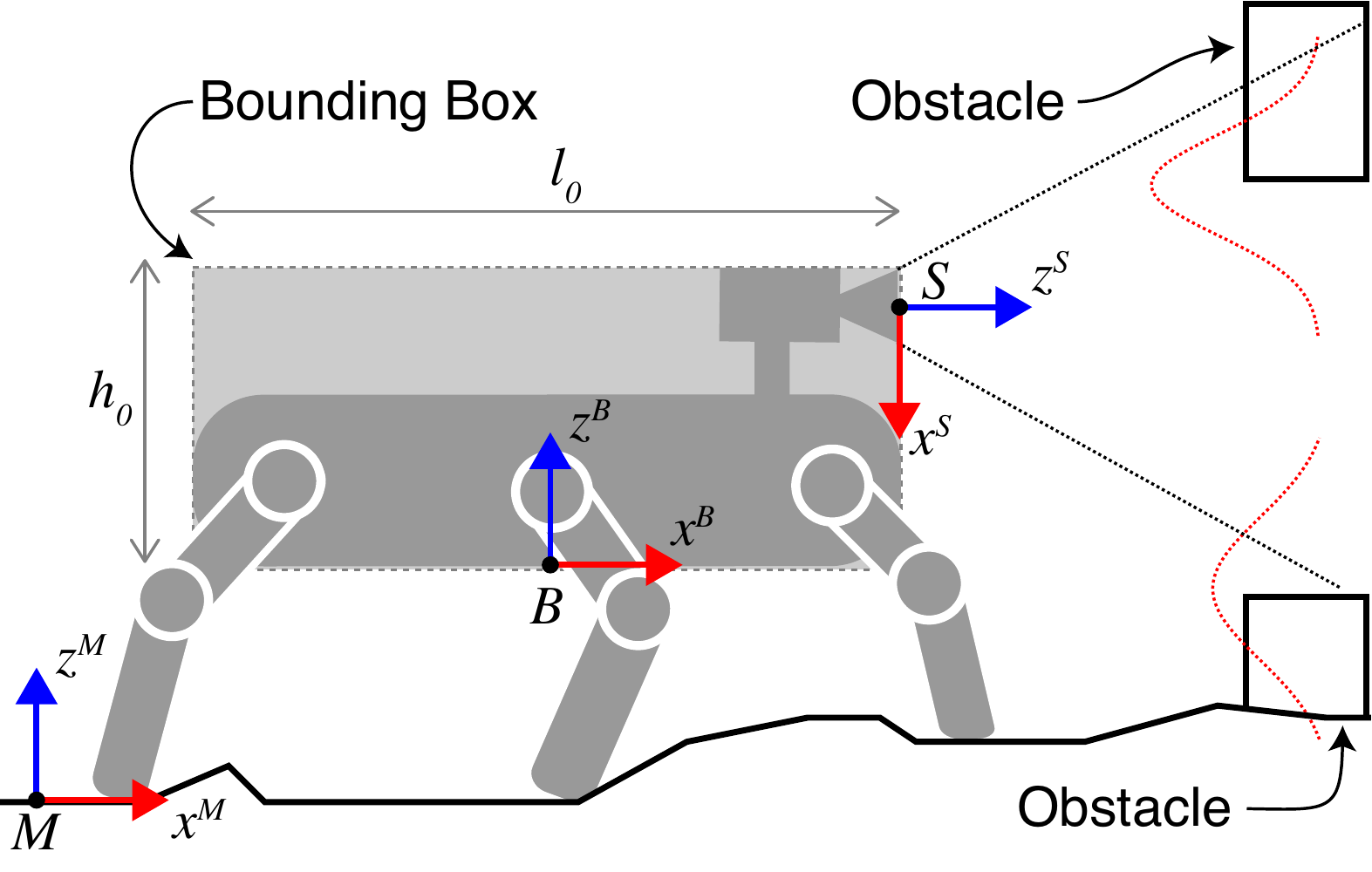}
    \caption{Illustration of a legged robot with bounding box, coordinate frames and elevation clustering. $M$, $B$ and $S$ origins indicate the map, body and sensor frames respectively. On the right of the figure is an obstacle observed by the sensor. The measurements are clustered into floor and ceiling elevations which are modelled as Gaussian distributions.
    }
    \label{fig:robot}
    \vspace{-0.5cm}
\end{figure}

Applying whole body contact planners to multilegged robots poses significant computational costs due to the high number of DOF. To plan robot body trajectories efficiently in 3D, we simplify this problem by introducing a deformable bounding box which encases the robot's body as shown in Fig.~\ref{fig:robot}. The box does not extend downwards to include the legs but does widen based on the robot's width. This simplification drastically speeds up collision checking but does not consider leg collisions. 

\subsection{Bounding Box Definition}
The bounding box has a fixed length and height but a variable width and is attached to the body frame B. We define $span$ $s$ as half the width of the bounding box so that the box dimensions are $[l_0,\, 2 \cdot s,\, h_0]$. The width extends to cover the widest points of the robot including extra width to account for the lateral component of leg swing.

\begin{figure}[b!]
    \centering
    \vspace{-0.5cm}
    \includegraphics[width=70mm]{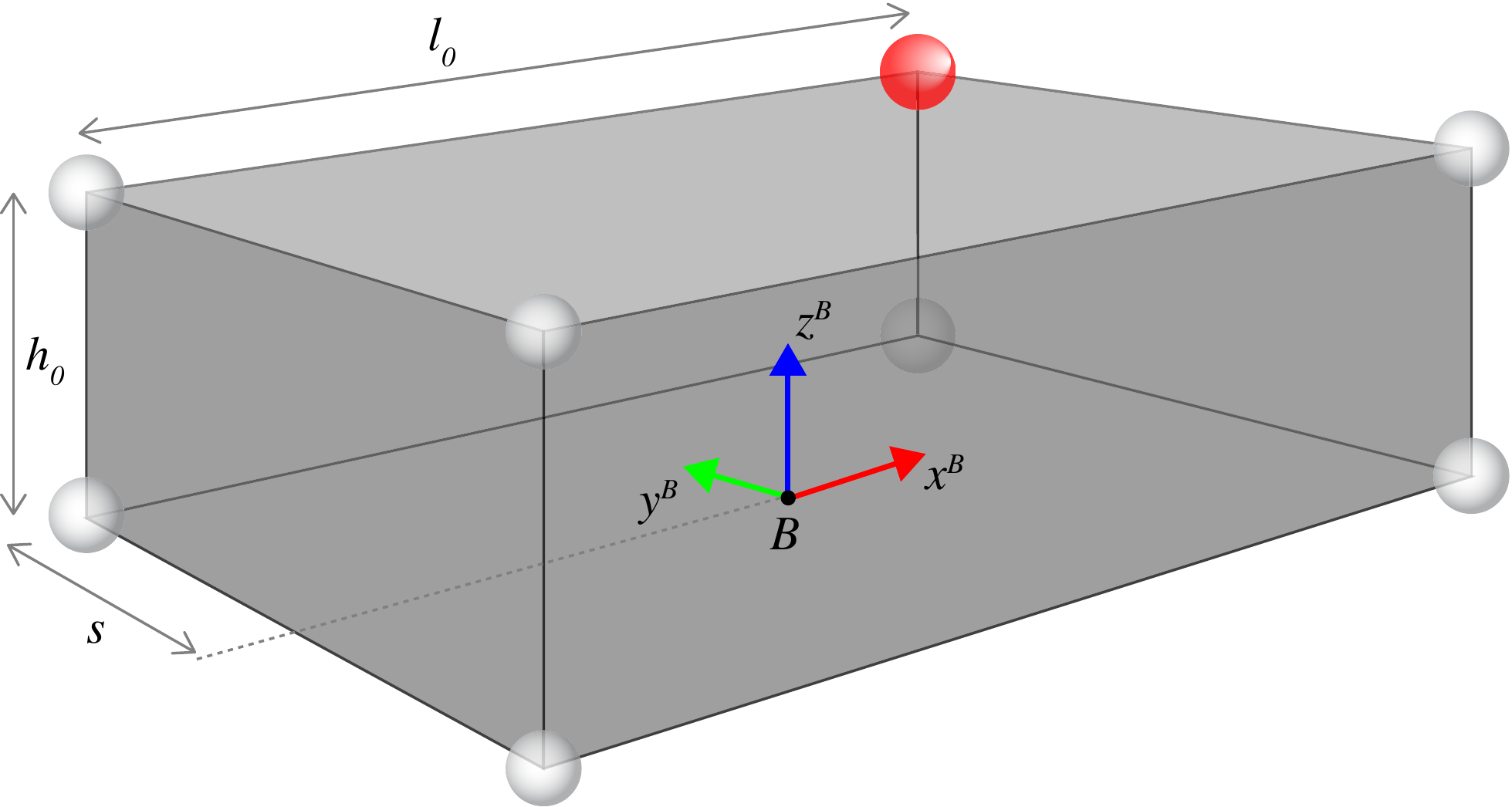}
    \caption{Deformable bounding box with some collision check points shown. The body frame is located at the centre of the bottom face. The red point has position $\mathbf{t}_{BC_j}^B = [\frac{1}{2} \cdot l_0,\ -1 \cdot s(z),\ h_0]^T$.}
    \label{fig:boundingBox}
\end{figure}

For collision checking $j$ points are defined around the bounding box. The locations of these points can be selected depending on the application or platform, Fig.~\ref{fig:boundingBox} shows the bounding box with a few points visualised. In our case, we define lines of points along each edge of the box. The number of points depends their radius which, at 5cm on Weaver, results in 84 points.

\subsection{Trajectory Definition}
\label{sec:trajectory}
We plan trajectories in 3D space from a start configuration $\xi(0)$ to a specified goal configuration with a series of configuration steps in time $\xi(t)$ given by 
\begin{equation}\label{eq:configuration}
    \xi(t) = [x(t), y(t), z(t), \phi(t), s(t)],
\end{equation}
where $x$, $y$ and $z$ are the position and $\phi$ the yaw of the robot base in the map frame $M$. We do not consider pitch and roll in this work. However, they could be included as well for more complex trajectories in future work. The trajectory also includes $s$ to deform the width of the bounding box as a function of time. From this point, we drop $t$ for notational simplicity.

The collision check points in our bounding box model can now be related to this trajectory definition with the vector
\begin{equation}\label{eq:collision_base_frame}
    \mathbf{t}_{BC_j}^B = [c_x \cdot l_0,\ c_y \cdot s(z),\ c_z \cdot h_0]^T.
\end{equation}
The subscript $BC_j$ indicates that the vector $\mathbf{t}$ is a translation from the $B$ frame to collision point $j$ and the superscript $B$ indicates that the vector is defined in the $B$ frame. Each element in $\mathbf{t}_{BC_j}$ is multiplied by a coefficient $c_{x,y,z}\in[-1, 1]$ which uniquely describes the collision point $j$ within the bounding box. For example a collision point located at the front, right and top vertex of the bounding box in Fig~\ref{fig:boundingBox} would have coefficients $c_x = 1, c_y = -1, c_z = 1$. 

Equation~(\ref{eq:collision_base_frame}) is expressed in frame $B$ and can be transformed to the map frame as:
\begin{equation}\label{eq:collision_map_frame}
    \mathbf{t}_{MC_j}^M(\xi) = \mathbf{t}^{M}_{MB}(x, y, z(s)) + \mathbf{R}_{BM}^M(\phi)\mathbf{t}_{BC_j}(s(z)),
\end{equation}
with the known translation $\mathbf{t}^{M}_{MB}(x, y, z)$ and yaw rotation of the robot $\mathbf{R}_{BM}^M(\phi)$. In (\ref{eq:collision_base_frame}), $s$ is a function of $z$ because we expect the width of the bounding box to deform as the robot raises or lowers its body. The converse is also a part of our model and we write $z(s)$ in (\ref{eq:collision_map_frame}).

 \begin{figure}[t]
     \centering
     \includegraphics[width=70mm]{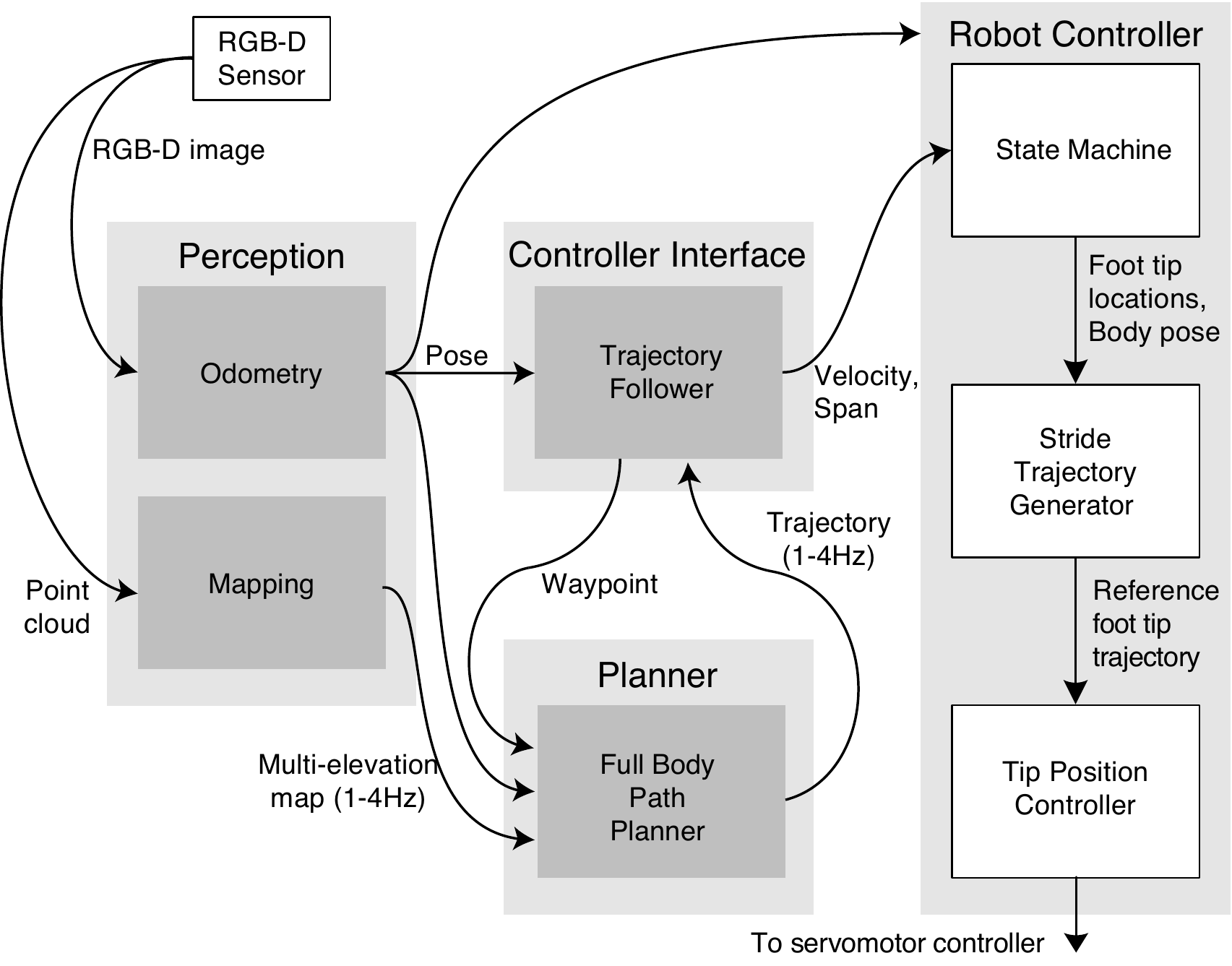}
     \caption{Overview of the system implemented on the robot.}
     \label{fig:architecture}
         \vspace{-0.5cm}
 \end{figure}

The position Jacobian of the collision check point $\mathbf{J}_{C_j} = \frac{\partial}{\partial \xi}\mathbf{t}_{MC_j}(\xi)\in \mathbb{R}^{3\times5}$ is given by
\begin{equation}\label{eq:jacobian}
\mathbf{J}_{C_j}  = 
\begin{bmatrix}
     \mathbf{I}_{3\times2},\
     \mathbf{R}\frac{\partial}{\partial z}\mathbf{t}_{BC_j},\
     \mathbf{R}'\mathbf{t}_{BC_j},\
     \mathbf{R}\frac{\partial}{\partial s}\mathbf{t}_{BC_j}
\end{bmatrix},
\end{equation}
with the following notation simplifications $\mathbf{R} = \mathbf{R}_{BM}(\phi)$ and $\mathbf{R}' = \frac{\partial}{\partial \phi}\mathbf{R}_{BM}(\phi)$.

The translation derivatives $\frac{\partial}{\partial z}\mathbf{t}_{BC_j}$ and $\frac{\partial}{\partial s}\mathbf{t}_{BC_j}$ relate how the span $s$ changes with $z$ as well as the converse. Unlike soft robots which have specific material properties, we can define this deformation relationship ourselves. For this work, we have chosen to model the relationship between $s$ and $z$ as linear, scaled between specified maximum and minimum limits for the robot's posture. However, one could use a more sophisticated model. Our choice has the benefit of being very simple but can potentially fail in some very narrow spaces, therefore our method is more conservative than other abstraction approaches.

\section{Navigation in Confined Spaces}
\label{sec:navigation}
Here we explain how to use our deformable abstraction to navigate through confined spaces. Figure~\ref{fig:architecture} shows an overview of the full implementation. We use an RGB-D depth sensor for both odometry and mapping of the local environment. 3D SDFs are generated to represent obstacles around the robot. We then plan optimised trajectories from the robot's current state to the requested waypoint. These trajectories are passed to a controller interface which follows the trajectory sending velocity and \textit{span} commands to the robot controller.

\subsection{Local Mapping}
\label{sec:local_mapping}
For fast 3D local mapping, there has been significant work recently for micro aerial vehicles (MAVs). Oleynikova et al.~\cite{oleynikova_voxblox:_2017} presented Voxblox which uses voxel hashing to very quickly build SDFs from distance measurements. Usenko et al.~\cite{usenko_real-time_2017} employ a robot-centric approach based on a 3D ring buffer to maintain an occupancy map around the robot. Both of these methods sacrifice accuracy for real-time performance and are typically used with 10\,cm or greater resolutions. This makes them less suitable for ground robots which operate very close to terrain and obstacles.

Elevation mapping has been successfully used for legged robots by storing detailed height information about the terrain \cite{herbert_terrain_1989}. To account for walls, elevation maps typically try to find the highest point of the terrain and therefore often map overhanging objects as part of the ground, hiding potential paths. Pfaff et al.~\cite{pfaff_efficient_2007} introduced one of the first multi-elevation maps which searched for overhanging obstacles to remove. This ensured the ground mapping was not affected by potential overhanging obstacles, however, it did not map the ceiling itself and it assumed the robot had a fixed height.

An additional concern for autonomous robots is the drift in odometry over time. Fankhauser et al.~\cite{fankhauser_robot-centric_2014} introduced robot-centric elevation mapping to address this issue by storing data from the robot's perspective and incorporating uncertainty from the robot's motion. The map is represented as a local 2D grid which moves with the robot, mapping new areas and discarding old, unreliable data, as the robot moves.

\subsection{Multi-Elevation Mapping}
\label{sec:mapping}
We extend the elevation mapping from \cite{fankhauser_robot-centric_2014} to map the \textit{ceiling} points above the robot as well as the \textit{floor} points below.  We do this using the Grid Map data structure \cite{Fankhauser2016GridMapLibrary} which allows multiple layers of 2D data to be stored in a grid centred on the robot as it moves. For each depth camera scan, the mean and variance $[\hat{h}, \hat{\sigma}^{2}_{h}]$ of the height measurement is updated in each cell by means of the following Kalman filter:
\begin{equation}\label{eq:kalman}
\begin{array}[t]{ll@{}ll}
\hat{h}^{+} = \frac{\sigma^{2}_{p}\hat{h}^{-} + \hat{\sigma}^{2-}_{h}\widetilde{p}}{\sigma^{2}_{p} + \hat{\sigma}^{2-}_{h}},
\end{array}
\quad 
\begin{array}[t]{ll@{}ll}
\hat{\sigma}^{2+}_{h} = \frac{\hat{\sigma^{2-}_{h}}\sigma^{2}_{p}}{\sigma^{2}_{p} + \hat{\sigma}^{2-}_{h}}
\end{array},
\end{equation}
where $-$ and $+$ superscripts indicate the filter states before and after a measurement respectively. The subscript $p$ indicates the unfiltered sensor measurement variance which comes from empirical models such as \cite{fankhauser_kinect_2015}. Before this fusion step, points are clustered into $floor$ and $ceiling$ elevations with means and variances $[\hat{h}_f, \hat{\sigma}_{f}^{2}]$ and $[\hat{h}_c, \hat{\sigma}_{c}^{2}]$. In Fig.~\ref{fig:robot}, on the right of the robot, these two elevation estimates are shown as Gaussian distributions.

 \begin{figure}[b!] \label{fig:clustering}
     \vspace{-0.3cm}
     \centering
     \includegraphics[width=70mm]{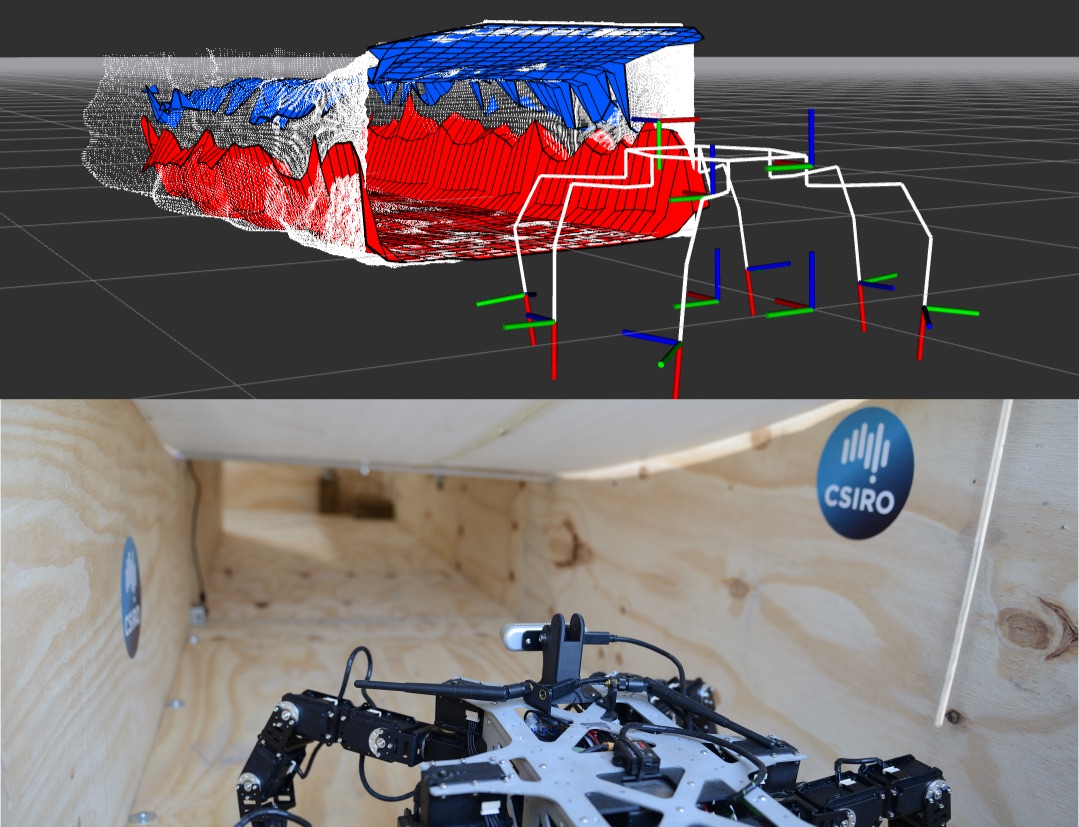}
     \caption{Hexapod robot Weaver in the test tunnel where the ceiling has been lowered to create and overhanging obstacle (bottom). A visual representation of the multi-elevation map with \textit{floor} (red) and \textit{ceiling} (blue) layers (top). The white points are the raw sensor measurements from the Intel Realsense D435 depth camera.}
 \end{figure}

Clustering is done by calculating the probability $P$ of a new point observation $\hat{h}$ belonging to an elevation $E$ which in our case is either a $floor$ or $ceiling$ elevation. Each of these elevations is modelled as $\mathcal{N}(\mu_E,\,\sigma_E)$ so that the probability likelihood function is given by 
\begin{equation}\label{eq:likelihood}
    P(\hat{h}_p|E) = \frac{1}{\sqrt{2\pi\sigma_{E}^{2}}}  e^{ -\frac{(\hat{h}_p-\mu_{E})^{2} } {2\sigma_{E}^{2}} }.
\end{equation}
In the case where there exists only one, or possibly no elevations, we use an elevation probability prior $P(E)$ which is based on the assumption that the floor will be below the robot's body and the ceiling above. This prior is a parameter which can be adjusted depending on how likely all \textit{floor} measurements are to be below the robot or \textit{ceiling} measurements above eg. a very low overhanging obstacle.

We then evaluate the posterior probability
\begin{equation}\label{eq:map}
    \hat{E}_{MAP}(\hat{h}_p) = \underset{E}{\arg\max} P(\hat{h}_p|E)P(E)
\end{equation}
for each elevation and classify the point observation based on the maximum \textit{a posteriori}. When there is only one elevation we use the height of the robot body as a reference to check if the measurement belongs to a new elevation. Once the point has been classified it can be fused into the corresponding elevation Kalman filter in (\ref{eq:kalman}).

Since clustering depends on the height of the robot's base, if an obstacle is below the robot, it becomes part of the floor and the robot can walk over it. If it is above the robot, it becomes part of the ceiling.

\subsection{Deformation Planning}
\label{sec:planning}
With the framework formalised in Section~\ref{sec:trajectory}, it would be possible to apply sampling methods as in \cite{grey_footstep} and \cite{orthey_quotient} by searching through the space of possible deformations. However, since our abstraction deformations are continuous and have well defined transitions via the Jacobians, we can also use trajectory optimisation methods. CHOMP is well suited to this problem as it is designed to be invariant to re-parameterisation and produces smooth trajectories.

Our trajectory optimisation uses the functional gradient descent update rule given by
\begin{equation}\label{eq:update}
    \xi_{i+1} = \xi_{i} - \frac{1}{\eta}A^{-1}\bar{\nabla}\mathcal{U}[\xi_i],
\end{equation}
where the norm $A$ is formed by multiplication of differencing matrices and acts as a smoothing operator on trajectories. The learning rate $\eta$ regulates the speed of convergence to a solution for each iteration $i$. $\bar{\nabla}\mathcal{U}$ is a functional gradient that operates on the trajectory configuration function $\xi(t)$ defined in (\ref{eq:configuration}). This functional gradient is the sum of two gradients $\bar{\nabla}\mathcal{F}_{smooth}$ and $\bar{\nabla}\mathcal{F}_{obstacle}$. $\bar{\nabla}\mathcal{F}_{smooth}$ is a cost on non-smooth trajectories calculated by
\begin{equation}
    \bar{\nabla}\mathcal{F}_{smooth}[\xi](t) = -\frac{d^2}{dt^2}\xi(t).
\end{equation}
Higher orders of derivative could be used as discussed in \cite{zucker_chomp:_2013}, however, we only use the 2nd order time derivative. The obstacle avoidance gradient $\bar{\nabla}\mathcal{F}_{obstacle}$ is calculated for each collision check point given by the sum
\begin{equation}
    \bar{\nabla}\mathcal{F}_{obstacle}[\xi] = \sum_{j=1}^{\mathcal{C}} \mathbf{J}_{C_j}^{T} \|{\mathbf{X'}}\|[(\mathbf{I} -\mathbf{ \hat{X}'}\mathbf{\hat{{X}'}}^T)\nabla c - c\kappa],
\end{equation}
where $\mathbf{J_{C_j}}$ is the position Jacobian for the collision check point as calculated in (\ref{eq:jacobian}). $\mathbf{X'}$ and $\mathbf{X''}$ are the velocity and acceleration for each collision point and $\mathbf{\hat{X}}$ denotes a normalised vector. Kappa $\kappa = \|\mathbf{X'}\|^{-2}(\mathbf{I}-\mathbf{\hat{X}'}\mathbf{\hat{{X}'}}^T)\mathbf{X''}$ is the curvature vector along the trajectory workspace. The matrix ${\|{\mathbf{X'}}\|[(\mathbf{I} - \mathbf{\hat{X}'}\mathbf{\hat{{X}'}}^T)}$ projects gradients orthogonally to the direction of motion to avoid affecting the speed profile. The variable $c$ represents the cost associated with a point in the trajectory being near an obstacle and comes from the SDF. The cost function used in this work is the same continuous piecewise function as proposed in \cite{zucker_chomp:_2013}.

\section{Experiments}
\label{sec:experiments}

In this section, we explain how we demonstrate the functionality of posture adaptation. We also  show that our method can solve similar problems as full joint space planners with significantly better computational performance.

\subsection{Experiment Tasks}
\begin{figure}[t!]
     \centering
     \includegraphics[width=70mm]{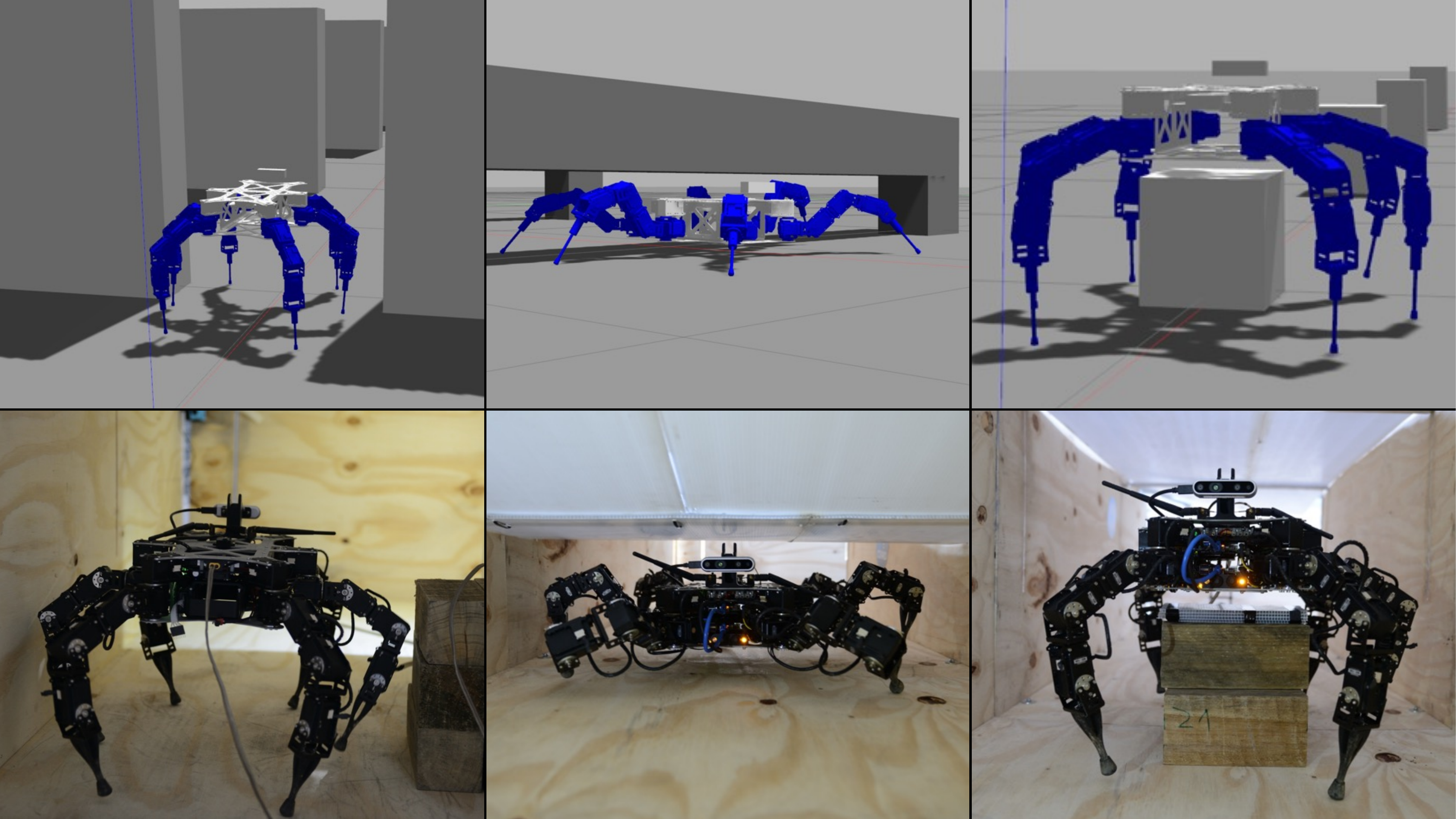}
     \caption{Three basic tasks attempted in simulation and on the real robot. From left to right: \textit{thin gap}, \textit{low overhang} and \textit{high clearance}.}
     \label{fig:tasks}
     \vspace{-0.5cm}
\end{figure}

We present three basic tasks: \textit{thin gap}, \textit{low overhang} and \textit{high clearance} which aim to show capability of our planning algorithm. Figure~\ref{fig:tasks} shows both the simulated robot and the real robot navigating each of these spaces. In the \textit{thin gap} task the robot must narrow its \textit{span} to reduce its width which, from our linear implementation of $\frac{\partial}{\partial s}\mathbf{t}_{BC_j}$, also results in a raised body. For \textit{low overhang} the robot must reduce its height which leads to a corresponding increase in width. \textit{High clearance} functions similarly in that the robot must raise its body so that the bounding box around the body passes over the obstacle. Since we do not plan the leg trajectories we set up this task so that the obstacle can always pass between the robot's legs.

 \begin{figure}[b!]
 \vspace{-0.5cm}
     \centering
     \includegraphics[width=70mm]{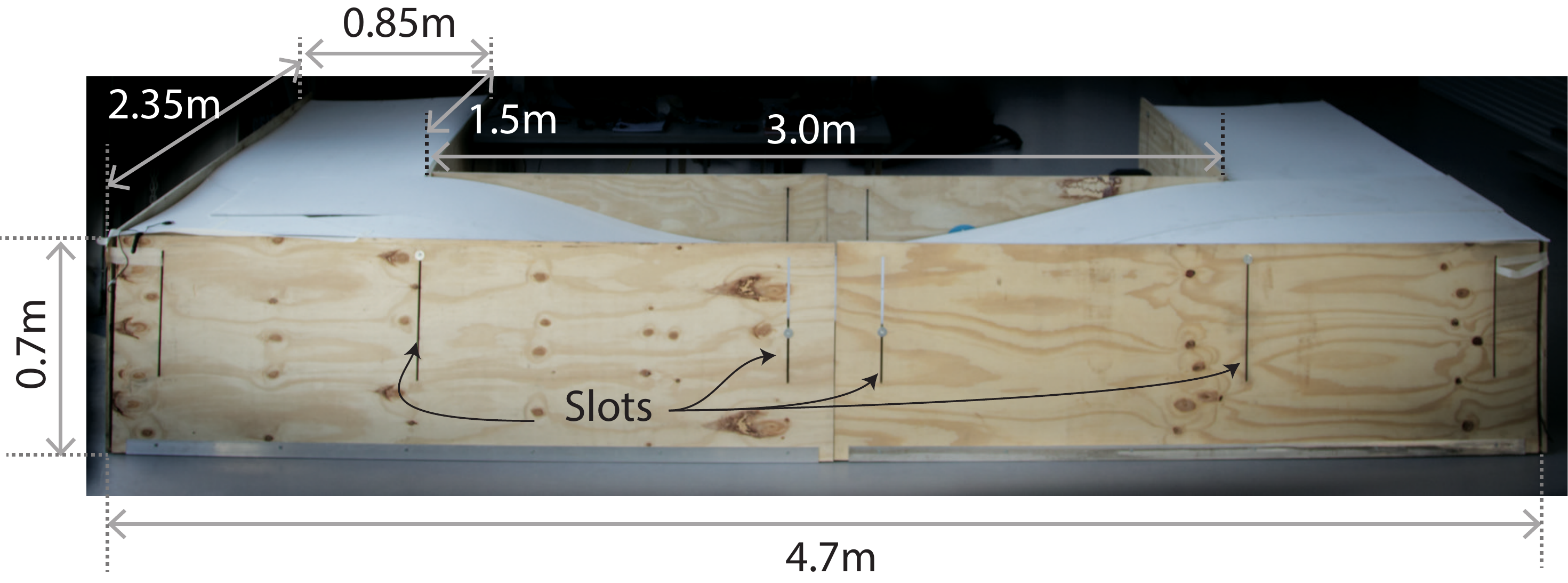}
     \caption{Test tunnel with slots for adjusting ceiling height.}
     \label{fig:testTunnel}
 \end{figure}
\subsection{Simulations}
The full navigation system in Fig \ref{fig:architecture} was implemented in a simulated environment. A hexapod robot, the environment and depth sensor measurements were all simulated. For each task, we experimented with progressively narrower spaces. To evaluate the amount of posture adaptation, we show a normalised percentage of displacement between a nominal value with which the robot normally walks and an empirically determined limit. For example, for the low overhang task, the simulated robot has a nominal height of 32.7\,cm and a minimum height of 14.1\,cm if the bottom of the body rests on the terrain. The total possible change in posture to pass under an obstacle is therefore 18.6\,cm. Thus lowering the body below 30\,cm corresponds to a displacement of 2.7\,cm or about 14.5\% of the total possible change in posture for this task. Using this posture adaptation percentage makes it possible to compare difficulties of tasks for different robot platforms.

\subsection{Performance Evaluation}
In the following we show that the presented planning framework is capable of navigating the same kinds of spaces as a whole body joint space planner, such as Contact Dynamic Roadmaps (CDRM)~\cite{short_legged_2018}. Moreover, we show that our framework greatly reduces the computation time. To demonstrate this benefit, we repeat their \textit{clearance} task and compare the performance with our framework.

For this task, a 15\,cm high block is progressively raised and the robot must walk forward to a waypoint 3\,m away. As the block is raised higher the robot first walks over the block then under. We tested with two robot models in simulation: the hexapod used in previous simulations and a quadruped robot model nearly identical to the one used in \cite{short_legged_2018}.

We measure the time taken to create the $6\times6\times1$\,m SDF from a given multi-elevation map and to then compute the desired trajectory. This already places our method at a disadvantage compared to CDRM which is given exact knowledge of the environment for their online planning step. In the original CDRM experiment, they used a quad-core i7 4700M CPU and 16\,GB of RAM; we could not find an exactly identical machine so for comparison we used a dual-core i7 5600U with 8\,GB of RAM which is a newer but significantly lower power machine. For an additional, more realistic comparison, we also recorded timing on the real robot's computer: an Intel NUC with a dual-core i7-5557U and 16\,GB of RAM.

 \begin{figure}[b!]
     \centering
     \vspace{-0.5cm}
     \includegraphics[width=70mm]{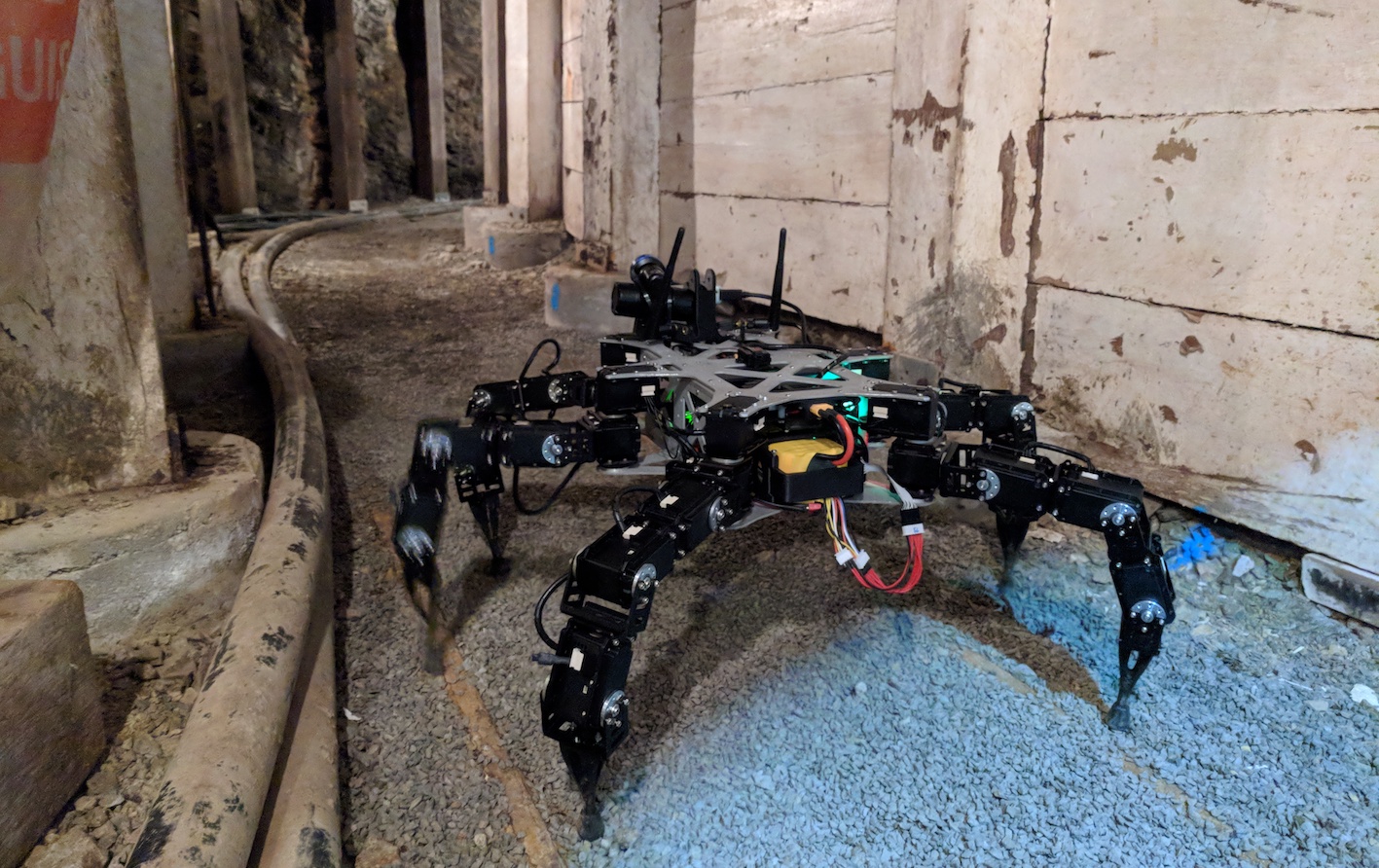}
     \caption{Weaver in a UQEM tunnel.}
     \label{fig:weaver_mine}
 \end{figure}
 
\subsection{Testing Tunnel}
In addition to simulations we implemented the full pipeline from Fig.~\ref{fig:architecture} on the hexapod robot Weaver~\cite{bjelonic_weaver:_nodate}. Weaver is a six-legged robot with 5 degrees of freedom for each leg and is capable of climbing 30\degree~inclines. Designed for proprioceptive sensing and adaptive based control, it is an ideal platform for demonstrating the capabilities of posture adaptation in realistic situations. For distance sensor, we use an Intel Realsense D435 to generate both the pointclouds and the RGB-D images used for odometry which is done with a modified version of ORB\_SLAM2 \cite{murORB2}.

Weaver uses a hierarchical whole body controller detailed in~\cite{bjelonic_weaver:_nodate}. When a deformation trajectory is generated, we pass the desired body pose for each step to the controller. We can also specify a maximum width of the robot and the controller computes desired footholds for the next step based on the specified gait.

We constructed an above-ground testing tunnel with adjustable ceiling that can be lowered to create overhanging obstacles (Fig.~\ref{fig:testTunnel}). Wooden blocks are used to create \textit{thin gaps} and \textit{high clearance} tasks inside the tunnel. Weaver accomplished each of these tasks inside the tunnel as shown in the bottom row of Fig.~\ref{fig:tasks}.

 \begin{figure}[t!]
     \centering
     \subfigure[High clearance task: Raise body above obstacles.]{
         \includegraphics[width=75mm]{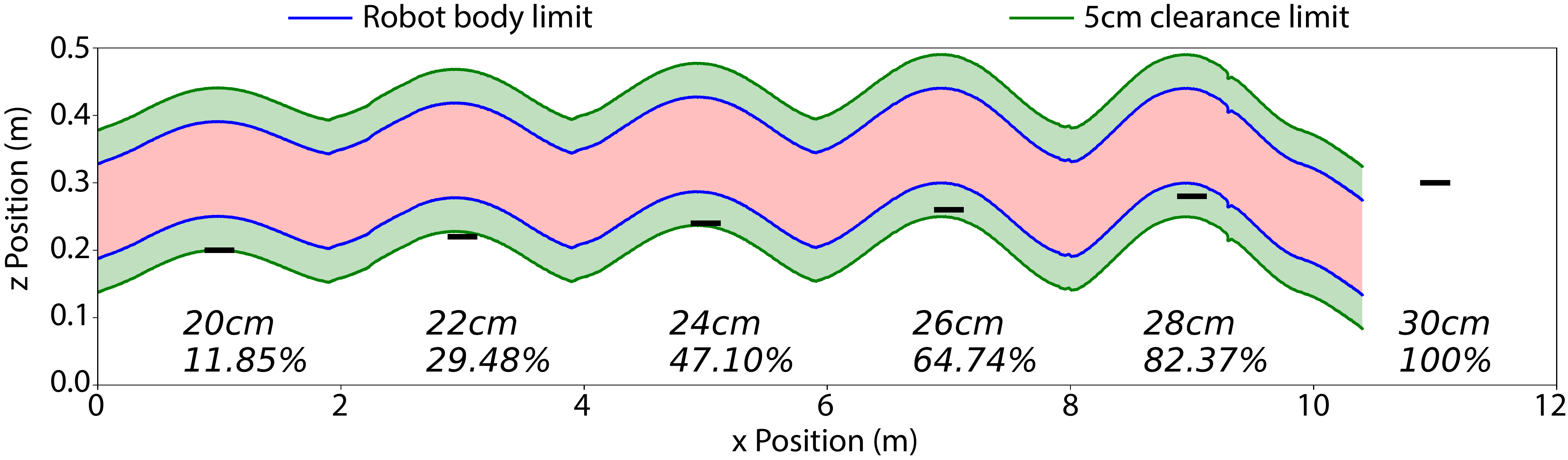}}
     \subfigure[Low overhang task: Crawl under an overhanging obstacle.]{
         \includegraphics[width=75mm]{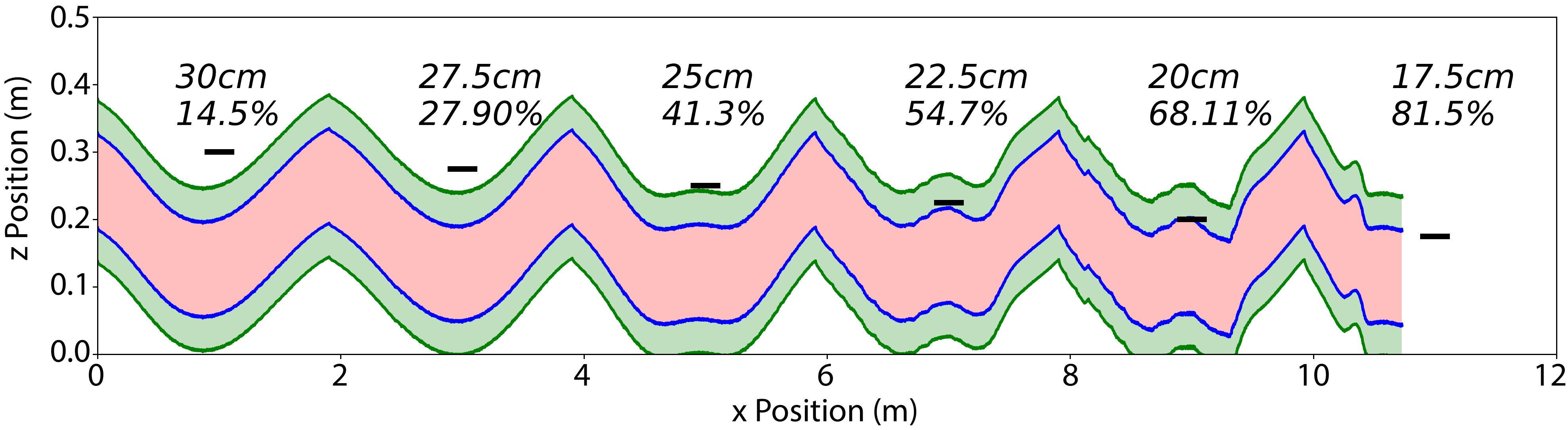}}
     \subfigure[Thin gap task: Reduce width (\textit{span}).]{
         \includegraphics[width=75mm]{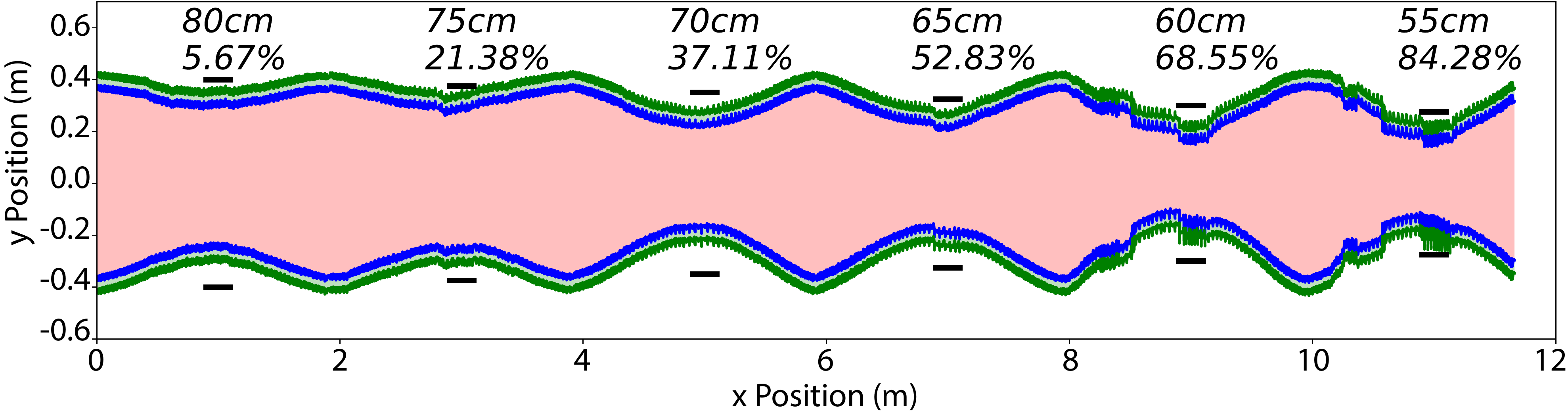}}
     \caption{Simulation results of the robot completing three tasks with progressively tightening constraints. Each plot is annotated with the constraint and the corresponding posture adaptation percentage.
     }
     \label{fig:all_sim}
     \vspace{-0.5cm}
 \end{figure}

\subsection{Field Experiments}
Weaver was also brought to the University of Queensland Experimental Mine (UQEM), a now-defunct silver and lead mine, to conduct field trials. The mine is administered by the University of Queensland's School of Mechanical and Mining Engineering. We brought the robot to a section of the mine where there are wooden supports on either side of the hallway with large concrete bases which make the hallway too narrow to normally navigate. Additionally, there is a large pipe running along one side of the hallway making the path even more narrow which can be seen in Fig.~\ref{fig:weaver_mine}. 

 \begin{figure}[b!]
    \vspace{-0.5cm}
     \centering
     \includegraphics[width=75mm]{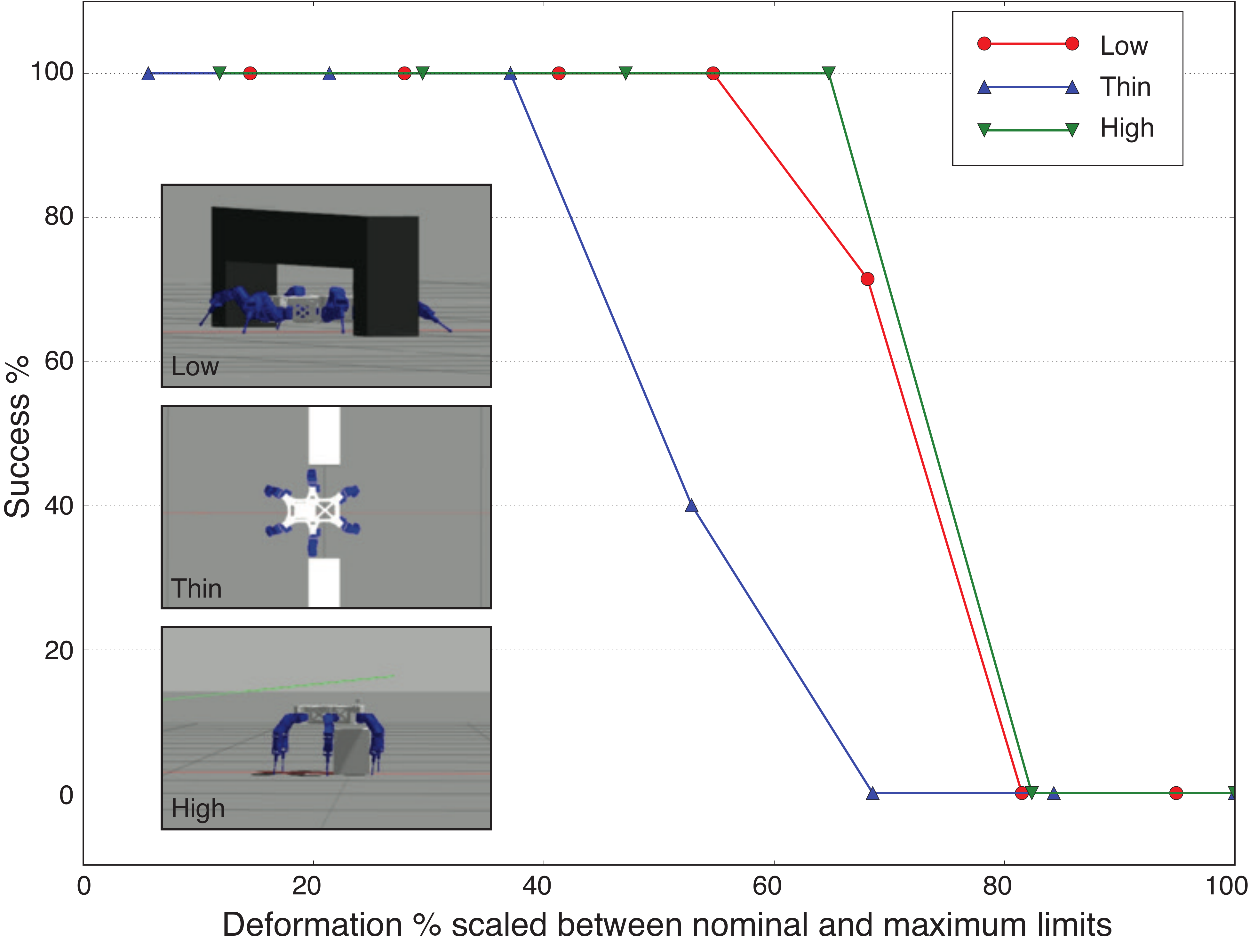} 
     \caption{Success rates of increasingly narrow/high spaces over 10 trials with 5\,cm collision radius (Insets: robot performing the tasks in simulation).}
     \label{fig:success}
\end{figure}

\section{Results and Discussion}
\label{sec:results}

Here we present our results and provide a discussion of the capabilities and limitations of our method.

\subsection{Simulations}
We found the robot was able to adapt its walking posture significantly and still navigate all of the tasks. Fig.~\ref{fig:all_sim} shows the simulated robot position as it traverses the same task with increasing difficulty. For all tasks the robot was able to plan posture changes of over 60\% and then carry them out, walking through the narrow space. Table \ref{table:deformations} shows the maximum posture adaptations that were capable without colliding with the environment. As shown in Fig.~\ref{fig:all_sim}, the robot was able to continue even further after a collision, passing through even more confined spaces.

In the \textit{high clearance} task, the robot walked over a 28\,cm high obstacle but the body collided with it. Since 30\,cm is the height limit for the simulated robot it was unable to plan any higher. This limitation can be seen in the Fig.~\ref{fig:all_sim} at 26\,cm and 28\,cm where the robot is forced to plan a trajectory with an obstacle in the green clearance area since it is prevented from raising its body any higher. 

\begin{table}[t!]
\centering
\caption {Minimum Confined Spaces with Collision-Free Navigation (with corresponding posture adaptation percentage).}
\label{table:deformations}
\begin{tabular}{cccc}
\toprule
  & High Clearance  & Low Overhang & Thin Gap \\
 \midrule
Simulation & 26\,cm (65\%) & 22.5\,cm (54.7\%) & 70\,cm (37.1\%) \\
Real robot & 22\,cm (65.1\%) & 25\,cm (53.0\%) & 70\,cm (37.1\%) \\
\bottomrule
\end{tabular}
    \vspace{-0.5cm}

\end{table}

For the \textit{low overhang} task, after 25\,cm the planner is also forced to generate trajectories with obstacles inside the clearance region. This is because the 5\,cm clearance effectively increases the total height $h_0$ of the robot body to be greater than 25\,cm making obstacle free trajectories are impossible. When this occurs CHOMP attempts to find the lowest obstacle cost trajectory by solving trajectories where the obstacles end up inside gaps between the collision check points. This can be mitigated by weighting $\bar{\nabla}\mathcal{F}_{smooth}$ higher than $\bar{\nabla}\mathcal{F}_{obstacle}$ but that could result in more collisions overall.

 \begin{figure}[b!]
         \vspace{-0.5cm}
     \centering
     \includegraphics[width=75mm]{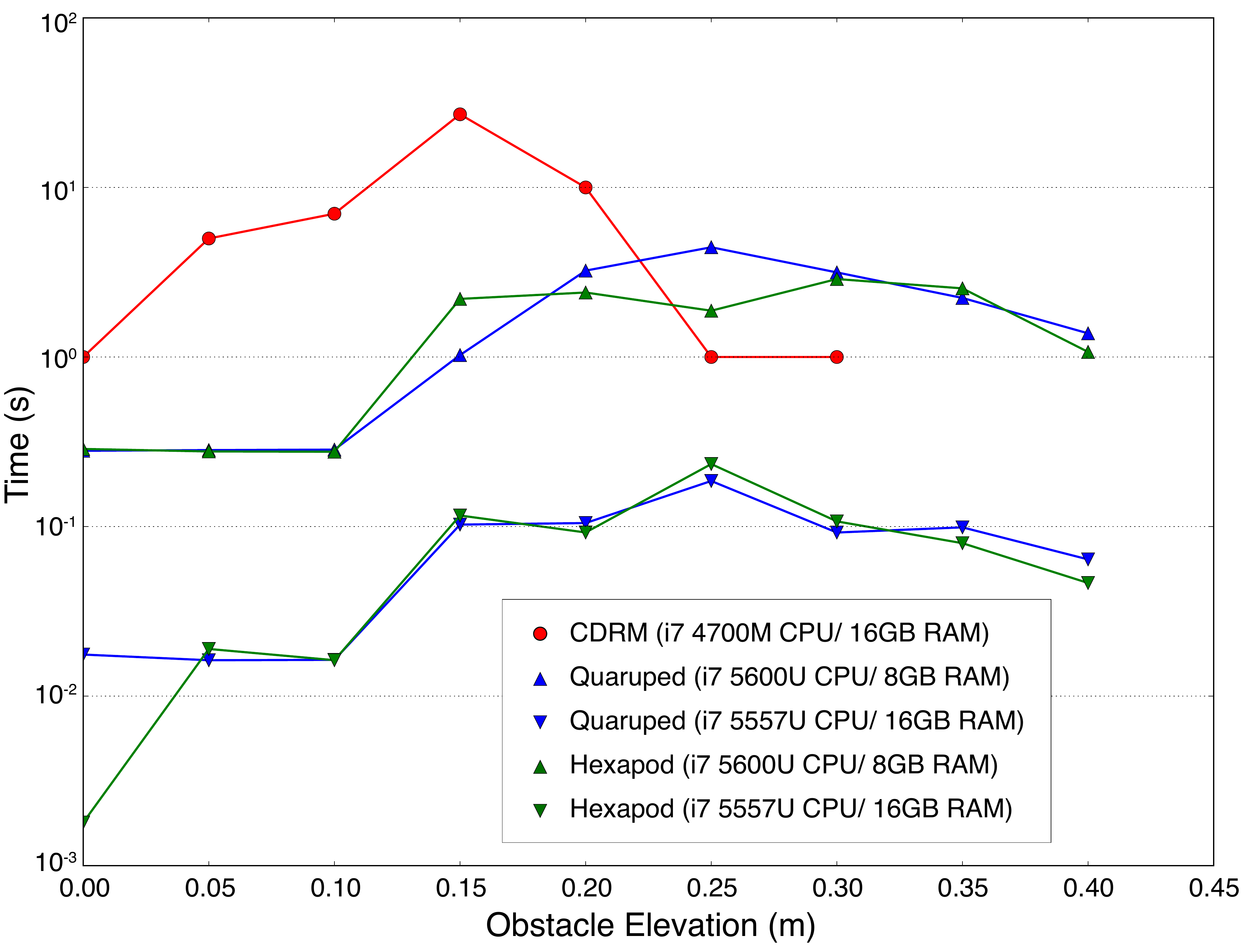}
     \caption{Comparison of CDRM planning time vs posture adaptation for the \textit{clearance} task on a log scale. Values for CDRM come from \cite{short_legged_2018}.
     }
     \label{fig:cdrmTIme}
\end{figure}

The most challenging task is the thin gap because our abstraction does not model the additional width from the robot taking a step. We account for this by adding a fixed span-offset to the robot. However, this has the drawback of artificially increasing the model's width which limits the amount of deformation for which the robot can plan.

Reducing \textit{span} also reduces the robot's support polygon, reducing stability which often results in collisions. We use a simple proportional controller to guide the robot along the generated trajectory so a more sophisticated control system might be able to account for this. \textit{High clearance} also has this instability effect however can achieve a higher percent posture adaptation simply because there is less potential for adaptation compared to \textit{span}: the simulated robot model can only raise its body 11.3\,cm but can potentially reduce its \textit{span} as much as 15.9\,cm.

We additionally examine the rate of failure of the planner and observe that the failure rate depends on the confinement of the environment. Fig.~\ref{fig:success} shows the results of 10 simulated trials for each task and each difficulty. In general, the robot is able to plan trajectories with 100\% success up to a certain confinement then, because of our conservative abstraction, the planner quickly fails.

\subsection{Performance Evaluation}
The framework is able to navigate the exact same environment and using a similar robot as CDRM. On the NUC (i7 5557U CPU / 16GB RAM), there is an improvement in performance of 1-2 orders of magnitude as shown in Fig.~\ref{fig:cdrmTIme} and all posture adaptation paths were planned in under 0.5\,s. This is possible because the deformable bounding box massively reduces the dimensionality of the problem from 18 (quadruped) and 36 (hexapod) to just 5. However, our method does not plan individual leg placement which prevents us from navigating other CDRM tasks. In Fig.~\ref{fig:cdrmTIme} there is a peak for all planning times where the planner must choose between going above or below the obstacle. In our experiments, this peak is slightly shifted which could be accounted for by differences in the robot models we used.

 \begin{figure}[t]
     \centering
     \subfigure[High clearance task: Raise body above obstacles.]{
         \includegraphics[width=75mm]{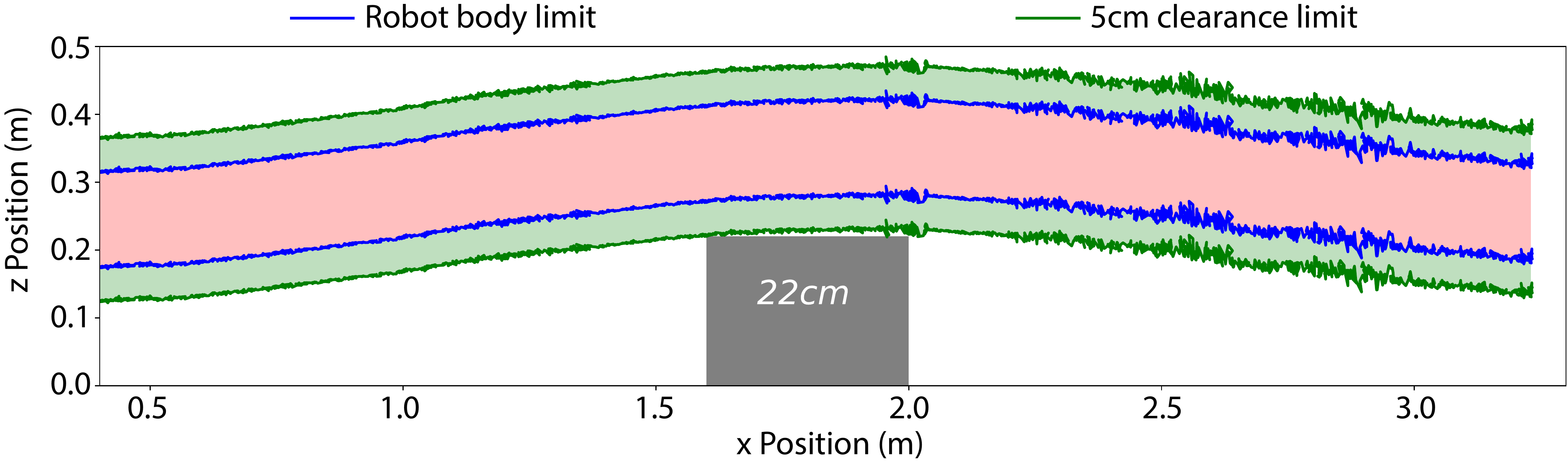}}
     \subfigure[Low overhang task: Crawl under an overhanging obstacle.]{
         \includegraphics[width=75mm]{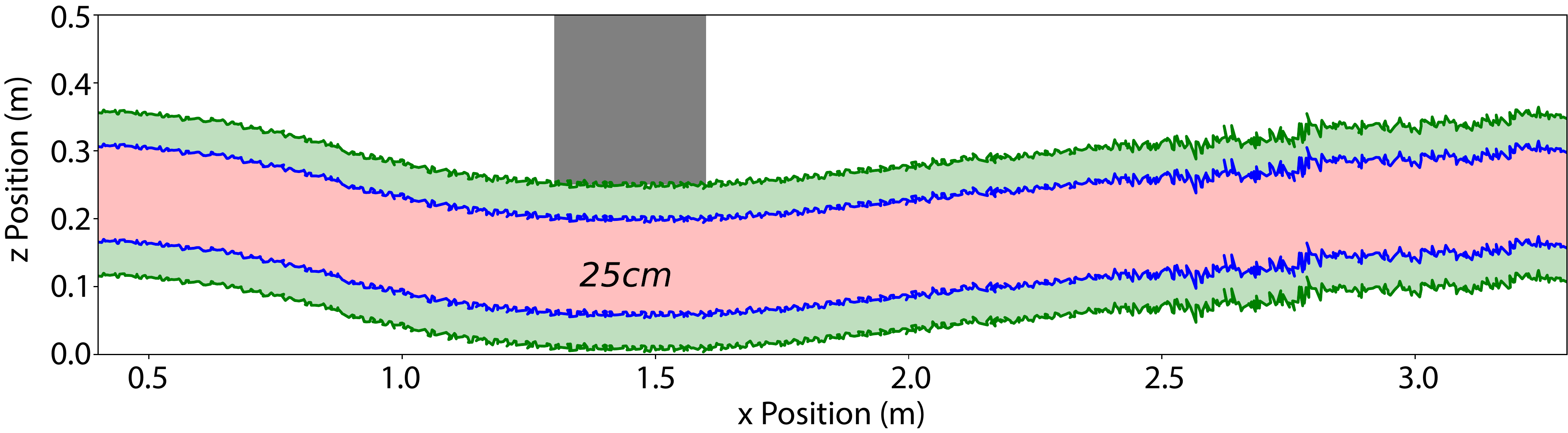}}
     \subfigure[Thin gap task: Reduce width (\textit{span}).]{
         \includegraphics[width=75mm]{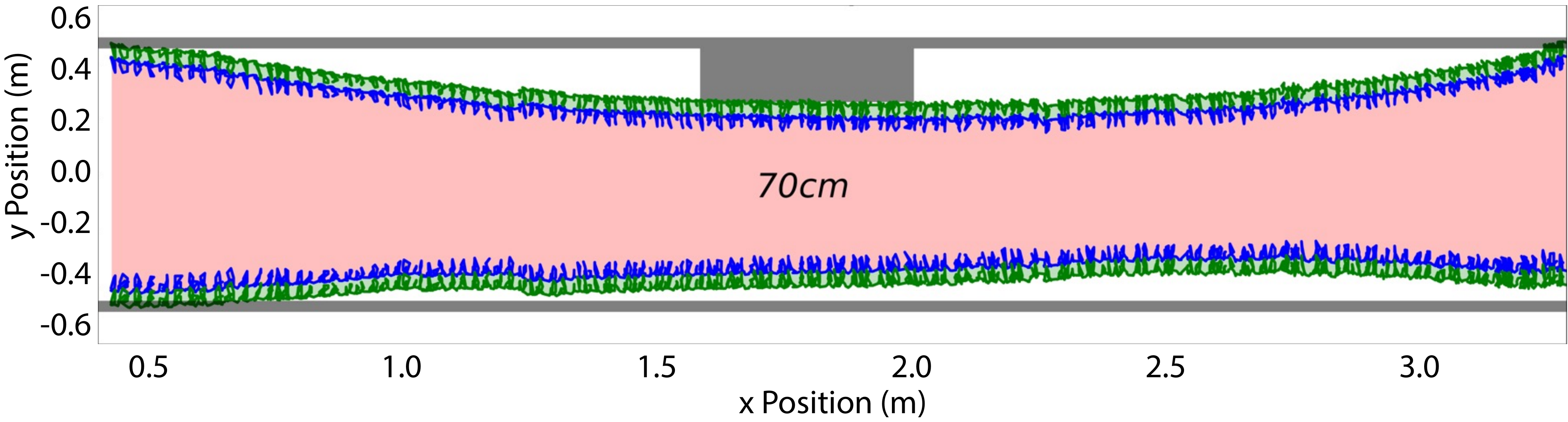}}
     \caption{Posture for each task on the real robot.
     }
     \label{fig:all_real}
     \vspace{-0.5cm}
 \end{figure}
 
\subsection{Testing Tunnel}
In Fig.~\ref{fig:all_real} we plot the position and width data recorded by the robot as it traversed the narrow spaces. Weaver was able to walk over a 22\,cm high obstacle, pass under a 25\,cm overhanging obstacle and reduce its width to 70\,cm without colliding with the environment. Table \ref{table:deformations} shows the corresponding percent posture adaptations. Note that the real robot is slightly different from the simulated model and stands over 33\,cm tall with a nominal clearance of 16.5\,cm which slightly changes the correspondence between confined space and posture adaptation percentage.

  \begin{figure}[t]
     \centering
     \includegraphics[width=75mm]{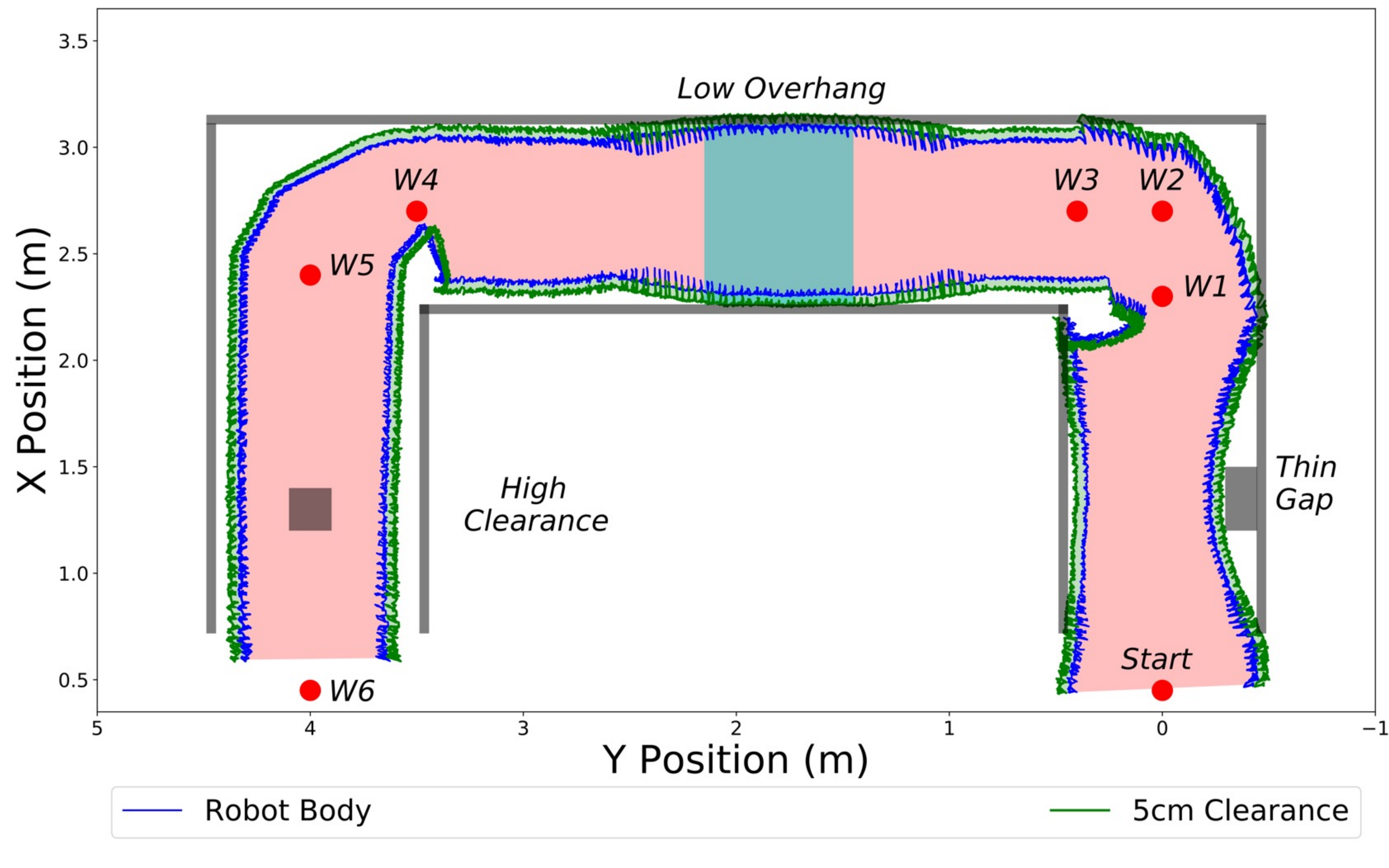}
     \caption{Results from completing the full obstacle course. Position information comes from the onboard odometry. The cyan rectangle indicates the area where there is a low overhanging obstacle. The red points indicate the waypoints sent to the path planner.
     }
     \label{fig:obstacle_course}
       \vspace{-0.5cm}

 \end{figure}
 
 \begin{figure}[b!]
         \vspace{-0.5cm}

     \centering
     \includegraphics[width=75mm]{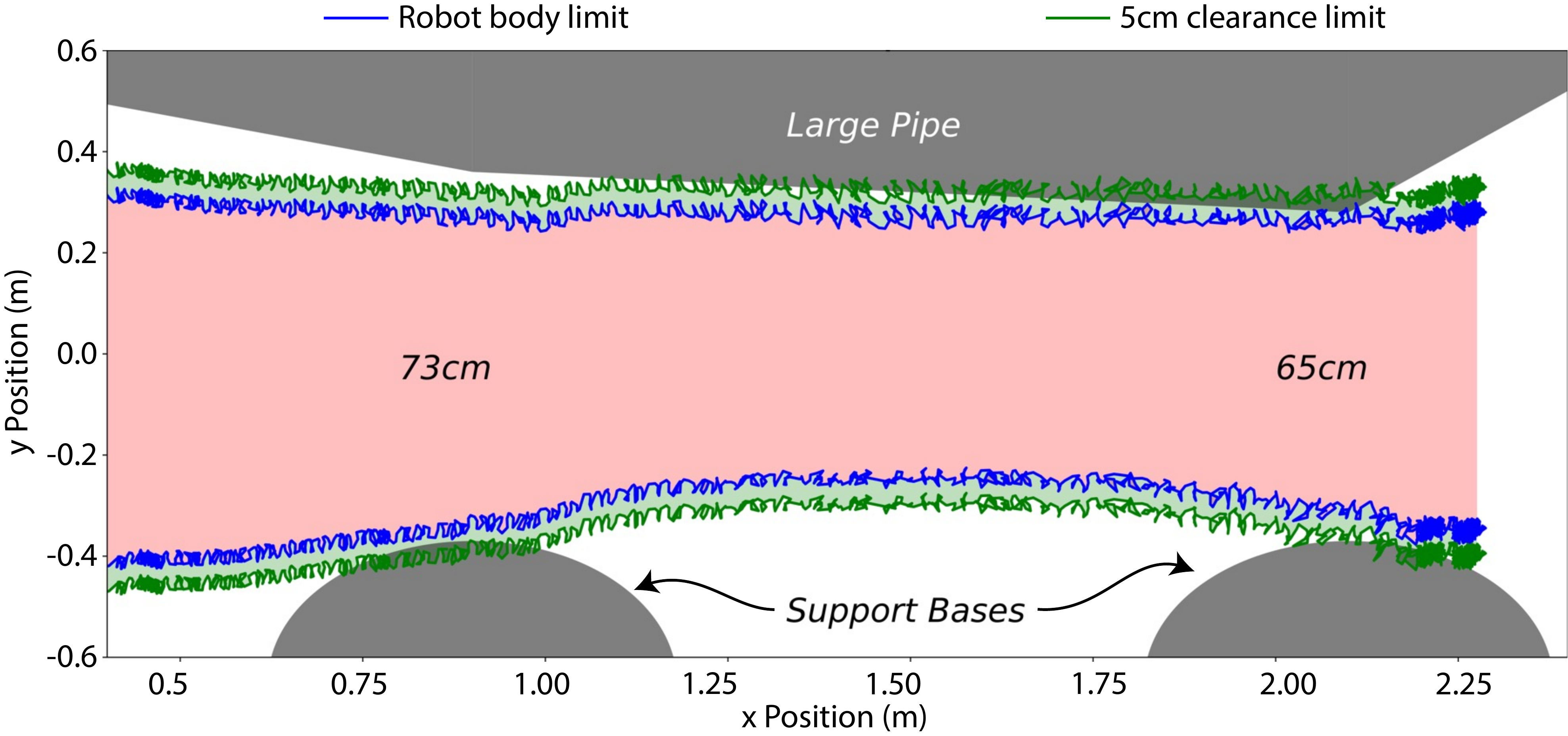}
     \caption{Results from field trial at UQEM showing the robot adapting its width to fit through the environment. Fig.~\ref{fig:weaver_mine} shows the the robot in this environment.
     }
     \label{fig:mine_data}
 \end{figure}
 
In addition to these tasks, we set up a complete obstacle course in which the robot must walk over 7.5\,m through the confined tunnel completing each of the tasks in sequence. Fig.~\ref{fig:obstacle_course} shows the path taken by the robot including the \textit{span} as it traversed the course.

In performing these experiments we experienced the same difficulties controlling the robot during the \textit{thin gap} task as seen in the simulations. Changing the walking gait from the usual tripod gait to a ripple gait helps with the instability. When attempting the complete obstacle course we experienced issues with obstacles being within the minimum specified range of the Intel Realsense. This led to ghost measurements above the robot which made the ceiling map very noisy.

\subsection{Field Experiments}
Fig.~\ref{fig:mine_data} shows a top-down view of the path and posture of the robot walking to a waypoint 3\,m away. Despite the space becoming as narrow as 65\,cm, which corresponds to a posture deformation of 52.83\%, the robot was able to reach the requested goal.

\section{Conclusion}
\label{sec:conclusion}
The objective of this work was to create a fast and reliable planner which allows legged robots to navigate confined spaces. This was achieved by using a deformable bounding box as an abstraction of the robot model. Moreover, this abstraction greatly simplified planning complexity in open and  confined spaces enabling us to solve challenging navigation problems efficiently. We additionally presented multi-elevation mapping of the local environment which was used to create SDFs of the space around the robot. While we deployed CHOMP to solve the deformation trajectories, any other planner could be used as our abstraction does not lose generality. We navigated confined spaces in simulation and on a real robot, showing feasibility. Finally, we performed field experiments in a real mine tunnel to fully demonstrate the usefulness and robustness of our proposed posture adaptation method. 

\section{Future Work}
Our planning framework is generic and light-weight enough that it can be combined with other, more sophisticated planners. One major goal for the future is to incorporate a leg swing planner such as in \cite{fankhauser_robust_2018} to completely avoid collision of the robot with the terrain. In that work they assume a constant body height and pre-generate footsteps based on an ideal gait pattern before optimising for terrain. It would be possible to instead use our body pose and width for the initial footstep generation. While this work can be directly applied to other robots, additional testing should be done with different abstractions and deformation models to find which solutions are best for different robot morphologies.

\section*{Acknowledgements}

The authors would like to thank Caio Fischer Silva, Benjamin Tam, Fletcher Talbot and James Brett for their assistance during this work, Brian White and Eric Muhling for facilitating experimentation at the UQEM.

\balance

\bibliographystyle{IEEEtran}

\end{document}